\documentclass[letterpaper, 10 pt, conference]{ieeeconf}  

\IEEEoverridecommandlockouts                              

\usepackage{graphics} 
\usepackage{epsfig} 
\usepackage{mathptmx} 
\usepackage{times} 
\usepackage[fleqn]{amsmath}
\usepackage{amssymb}  
\usepackage{booktabs}
\usepackage{algorithm}
\usepackage{algorithmic}
\usepackage[switch]{lineno}
\usepackage{svg}
\usepackage{soul}
\usepackage{url}
\usepackage[hidelinks]{hyperref}
\usepackage[utf8]{inputenc}
\usepackage[small]{caption}
\usepackage{graphicx}

\newcommand{\synesthesia}{$S$-FRIDA}

\title{\LARGE \bf
Robot Synesthesia: A Sound and Emotion Guided Robot Painter
}

\author{Vihaan Misra$^{1}$, Peter Schaldenbrand$^{1}$, and Jean Oh$^{1}$
    \thanks{
        $^{1}$The Robotics Institute, Carnegie Mellon University
    }
    \thanks{
        \{vihaanm, pschalde, hyaejino\}@andrew.cmu.edu
    }
}

\begin{document}

\maketitle
\thispagestyle{empty}
\pagestyle{empty}

\begin{abstract}

If a picture paints a thousand words, sound may voice a million.  
While recent robotic painting and image synthesis methods have achieved progress in generating visuals from text inputs, the translation of sound into images is vastly unexplored. 
Generally, sound-based interfaces and sonic interactions have the potential to expand accessibility and control for the user and provide a means to convey complex emotions and the dynamic aspects of the real world. 
In this paper, we propose an approach for using sound and speech to guide a robotic painting process, known here as \textit{robot synesthesia}. For general sound, we encode the simulated paintings and input sounds into the same latent space. For speech, we decouple speech audio into its transcribed text and the tone of the speech. Whereas we use the text to control the content, we estimate the emotions from the tone to guide the mood of the painting.
Our approach has been fully integrated with FRIDA, a robotic painting framework, adding sound and speech to FRIDA's existing input modalities such as text and style. 
In two surveys, participants were able to correctly guess the emotion or natural sound used to generate a given painting more than twice as likely as random chance. 
On our sound-guided image manipulation and music-guided paintings, we discuss the results qualitatively. 

\end{abstract}


\section{Introduction}

FRIDA~\cite{frida2022,schaldenbrand2024cofrida} is a collaborative robotic framework for creating real-world paintings. FRIDA can produce an artistic visualisation that is consistent with the human artist's intent expressed in the text form. 
Whereas text description in language can succinctly express semantic concepts in an abstract manner, it can be challenging to fully express the complexity and nuances of sounds or emotions, e.g., people use emojis to express their emotions in textual communication. 

Art has traditionally served as a powerful medium for expressing and evoking human feelings and emotions. 
Emotions form an integral part of the painting process, and it is crucial to understand the relationship between visual art and emotions to capture the painter's actual intent. 
Studies find that different visual elements such as color, shape, and composition can evoke different emotions in viewers~\cite{color_emo_relationship,achlioptas2021artemis}.
By focusing on sounds and the emotions behind it, we aim to initiate a more nuanced perceptual understanding of the painting process, which, downstream, can also be applied to richer understanding of ordinary paintings. 
%
Additionally, a speech interface can improve the usability and accessibility of the system for users with motor, linguistic, and cognitive impairment~\cite{accessibility2020}.

\begin{figure}
    \centering \small
    \includegraphics[width=\columnwidth]{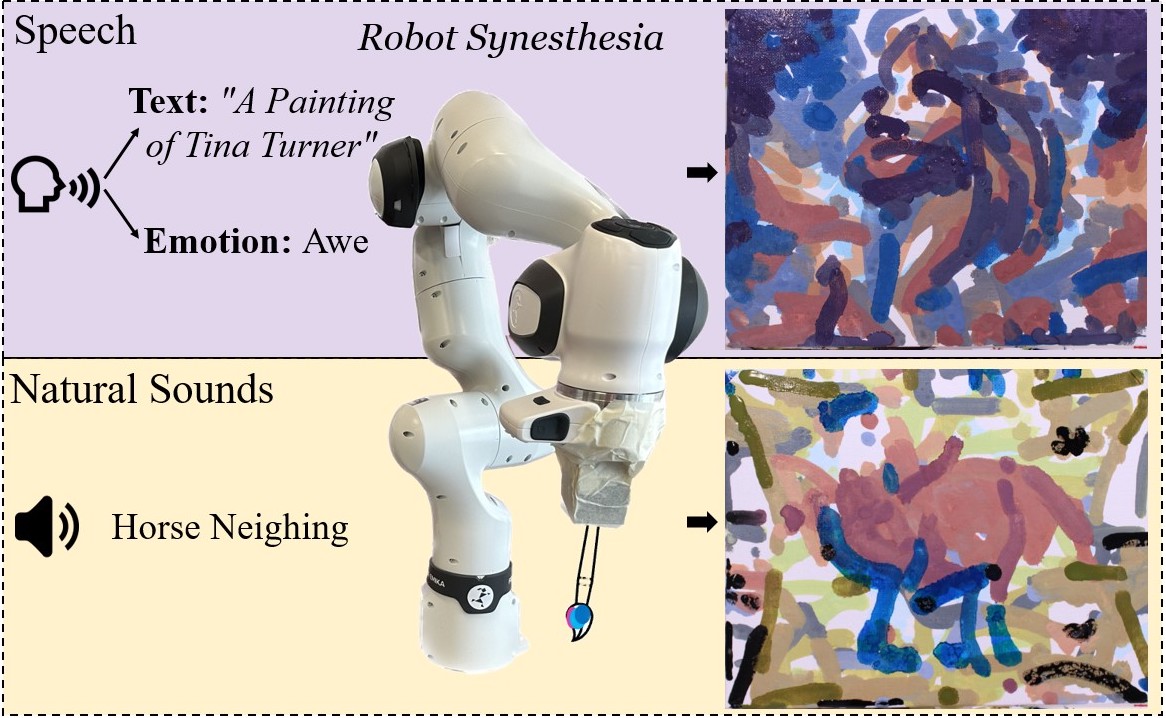}%
    \caption{
    \textbf{Robot Synesthesia} connects a robot painter's action space directly to user driven sonic interactions. For \textit{Speech guidance}, sound is decoupled into language and emotion whereas with \textit{Natural sound guidance}, the audio itself drives the content of the painting.
    }
    \vspace{-10pt}%
    \label{fig:main_fig}%
\end{figure}

Toward intuitive and powerful painting generation, we propose an approach to use the sound and speech input modalities for painting, introducing  \textit{Synesthesia} FRIDA (\synesthesia). Synesthesia is a perceptual phenomenon in which a person may perceive visuals when listening to sounds.

\begin{figure*}[ht]
    \centering \small
    \includegraphics[width=\textwidth]{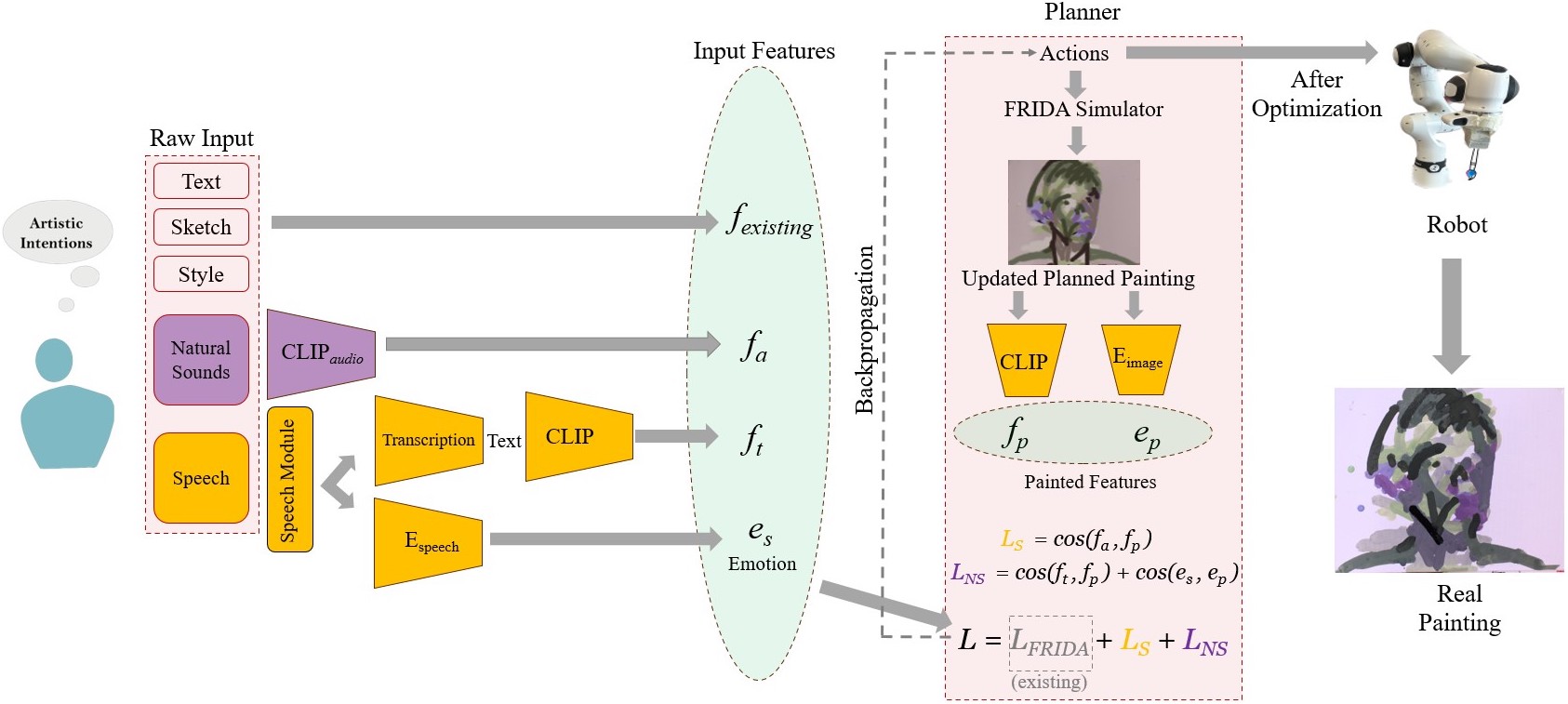}\vspace{-7pt}
    \caption{
    \textbf{\synesthesia{}  Overview } A human user's artistic intentions are specified via any combination of natural sounds or speech as well as any of FRIDA's~\cite{frida2022} existing input modalities, e.g., style or sketches. Features are extracted from the audio, or in the case of speech, a transcription and emotions are estimated. Brush stroke actions are rendered into a simulated painting using FRIDA, then features are extracted and compared to the input features to form loss functions. The loss is backpropagtated and gradient descent updates the actions to decrease the loss. After optimization, the brush stroke actions are executed by a robotic arm (UFactory XArm and Franka Emika platforms). We observe a high degree of fidelity between the simulated and the real painting drawn by the \synesthesia{} framework.
    }
    \label{fig:approach}
\end{figure*}

In contrast to powerful text-image models, existing audio-image models tend to focus on limited types of natural sounds~\cite{sound_guided_cvpr,wav2clip}, lacking semantic and emotional context understanding. To overcome this limitation, we treat speech as a special case of sounds differentiated from natural sounds by having content specified in language with an emotional context. We incorporate transcription and speech emotion recognition modules to enhance the understanding of the audio input. 
The transcribed text is processed the same manner as the text input in the original FRIDA system where it is encoded into the input latent space. The differences between the input features and the features of the planned painting form the loss function for the planner to decide the brush stroke actions. To recognize emotion from the input audio, we use speech emotion recognition (SER)~\cite{burnwal2020-speechEmotionKaggle} trained with an additional dataset for robustness. We use a pretrained image2emotion model~\cite{szymczak2022emotionPrediction} to estimate emotion from the current canvas and then compare that with the emotion recognized from the input audio to form an additional loss term for the planner. 

While other sound-guided image manipulation methods are constrained by the dataset used to train their generator model~\cite{sound_guided_cvpr}, \synesthesia{} does not rely on a particular pretrained generator and can, therefore, use sound guidance with arbitrary images.

Our surveys show promising results on audio guided paintings, across both natural sounds and emotions as modalities. Furthermore, we qualitatively demonstrate the application of these results through a series of simulated as well as real-world paintings.

Our contributions are as follows:
\begin{enumerate}
\item We developed a method for sound-guided image manipulation that generalizes across different content types in the painting domain, offering a novel way to integrate auditory cues into visual artwork creation.
\item We crafted a speech-to-painting approach that separates speech input into content--interpreted through text--and mood--understood via speech emotion. This method aims to capture both the literal and emotional essence of the spoken input in the artwork.
\item To support further research and collaboration, we will make our code publicly available upon the acceptance of this work, hoping to contribute to the broader community's efforts in this area.
\end{enumerate}
These efforts are intended to provide foundational steps toward leveraging sound and speech as well as emotions as inputs for robotic painting, aiming to capture a richer palette of human expression in art created by robots.

\section{Related Work}

\subsection{Sound-Image Encoding}

Encoding sounds and images into the same latent space in a differentiable manner allows for comparison of the two modalities that can be exploited such that one of the modalities can be altered to match the other. \cite{sound_guided_cvpr} train a sound encoder, $CLIP_{Audio}$, to encode sound and images into the same latent space. $CLIP_{audio}$ was trained on the VGG-Sound Dataset~\cite{chen2020vggsound} which contains over 200k clips for 309 different classes.  The dataset contains 200,000 10-second audio clips from each class captured from YouTube videos, with no more than two clips per video. The dataset's sound categories can be broadly divided into: people, animals, music, sports, nature, vehicles, homes, and tools, among others. While $CLIP_{Audio}$ performs well for simple natural or abstract sounds such as the sound of crying or thunderstorms, it was not designed to understand spoken language and its semantic meanings. 

Building on the concept of multimodal learning, Wu et al.~\cite{wav2clip} introduced Wav2CLIP, a method that distills robust audio representations from the Contrastive Language-Image Pre-training (CLIP) model. By leveraging the visual model of CLIP, Wav2CLIP trains an audio encoder to project audio into a shared embedding space with images and text, facilitating direct comparison and retrieval across modalities. 

\subsection{Emotion-Guided Image Generation}
Although CLIP based audio-visual models show good results in image generation, few existing works have studied emotions.  \cite{ibarrola2023affectImageGen} used a trained neural network as a feedback function, similar to \synesthesia{}, however, the dataset the model was trained on was only 1000 images, none of which were from an artistic domain.

\subsection{Sound-Guided Visual Manipulation}

Many previous works have added sound-guidance to existing image generators to perform sound-guided image or video manipulation.  Soundini \cite{lee2023soundini} uses sound to edit videos, and MM-Diffusion~\cite{ruan2023mmdiffusion} generates audio-video simultaneously.
The latent space of pre-trained image generators, such as Style-GAN, has been manipulated to understand sound as input~\cite{sound_guided_cvpr, styleclip_2021}.
In other works, music can guide the image generator~\cite{traumer_2021} or perform style transfer~\cite{lee_crossing_in_style_2020}.

While there has been success in using this methodology for image manipulation, previous work fails to create generalizable output images fitting the sound descriptions because they utilize pre-trained image synthesis models that can only generate images within their training distribution such as churches, faces, or artwork. Therefore, these prior works use sound only to manipulate a particular image that the synthesis model is capable of producing. In this work, our approach creates in a more general content domain and can generate images purely from audio inputs.

\section{Background}
 The proposed approach adds direct speech, sound, and emotional guidance to the FRIDA robotic painting system~\cite{frida2022}.  Brush strokes form the action space, in FRIDA, which are parameterized by values representing stroke length, bend, and thickness as well as position, orientation, and paint color. Given a set of brush stroke actions and a picture of the canvas, FRIDA's simulation environment can differentiably render the strokes and layer them onto the canvas image forming  a simulated painting represented as an RGB image.  
 A loss value can be computed using the simulated painting, and due to the differentiability of FRIDA's simulation, gradient descent can be used to optimize the brush stroke parameters to minimize the loss.

\section{Approach}
 
 Our approach, depicted in Fig.~\ref{fig:approach}, adds sound input to the FRIDA system. Because sounds have vastly different categories, we create approaches for two categories of sound: 1) natural sounds, which are diverse sound samples from various sources, and 2) speech sounds, which are a specific subset of natural sounds made by humans composed of language and tone.

 \subsection{Natural Sounds Guidance}

 For natural sound guidance, we use the pretrained $CLIP_{audio}$~\cite{sound_guided_cvpr} audio-image encoder trained on a wide variety of labelled sounds from YouTube videos.
 We encode the simulated painting, $p$, and input audio, $a$, into the same latent space using $CLIP$~\cite{radford2021-clip} and $CLIP_{audio}$ respectively. The encodings are compared using cosine distance to form a loss function. In practice, the painting is augmented using various perspective warps and cropping, as is customary in CLIP-guided image synthesis, for robust loss backpropagation.

\begin{equation}
 \begin{aligned}
 f_a &= CLIP_{Audio}(a); &f_p = CLIP(p)
\end{aligned}
\end{equation}\vspace{-0.5cm}
\begin{equation}
 \begin{aligned}
L_{\text {NS }}(p, a) &= \cos ( f_p,  f_a)
\end{aligned}
\end{equation}


\subsection{Speech Guidance}

Speech is an special subset of sound that humans use to express their ideas and feelings. We focus on two distinct features of speech, i.e., language and tone to represent the content and its additional emotional context, respectively. Our approach to speech guided painting decouples speech into text and emotions. 
While $CLIP_{Audio}$ is a promising tool for natural sounds, we find that it is not suitable for comprehending rich semantic meanings of spoken language. 
To utilize the semantic meaning of the text input, we leverage the text-image $CLIP$ instead and interpret the audio input separately for the moods. 

The input speech is transcribed to text, $t$, using the Whisper~\cite{radford2022Whisper} model.  This text can then guide the painting using FRIDA's existing text-to-painting methodology which encodes both the text and simulated painting into the same latent space using CLIP  ($f_t$ and $f_p$, respectively).
The cosine distance in features forms a loss function as seen in the first term of Eq.~\ref{eq:loss_speech}.

\begin{equation}
 \begin{aligned}
 f_t &= CLIP(t); &f_p = CLIP(p) \\
 e_p &= E_{img}(p); &e_s = E_{speech}(a)
\end{aligned}
\label{eq:feat_extract}
\end{equation}\vspace{-0.3cm}
\begin{equation}
 \begin{aligned}
L_{\text {S }}(p, a) &= \cos ( f_p,  f_t) + \cos(e_s, e_p)
\end{aligned}
\label{eq:loss_speech}
\end{equation}

\begin{table}[]
\footnotesize
\begin{tabular}{@{}llllll@{}}
\toprule
ArtEmis                                                  & Ravdess  & Crema & TESS     & SAV & IEMOCAP                                                  \\ \midrule
amusement                                                & happy    & HAP   & happy    &     & hap, exc                                                 \\
anger                                                    & angry    & ANG   & angry    & a   & fru, ang                                                 \\
awe                                                      &          &       &          &     &                                                          \\
contentment                                              & calm     &       &          &     &                                                          \\
disgust                                                  & disgust  & DIS   & disgust  & d   & dis                                                      \\
excitement                                               & surprise &       & surprise & su  & sur                                                      \\
fear                                                     & fear     & FEA   & fear     & f   & fea                                                      \\
sadness                                                  & sad      & SAD   & sad      & sa  & sad                                                      \\
\begin{tabular}[c]{@{}l@{}}something\\ else\end{tabular} & neutral  & NEU   & neutral  & n   & \begin{tabular}[c]{@{}l@{}}neu, xxx, \\ oth\end{tabular} \\ \bottomrule
\end{tabular}
\caption{Correspondence between emotion labels in different datasets. From left to right: ~\protect\cite{achlioptas2021artemis}\protect\cite{livingstone2018ravdess}\protect\cite{cao2014cremaD}\protect\cite{dupuis2010tess}\protect\cite{jackson2014surrey}\protect\cite{busso2008iemocap}}
\label{tab:emotions}
\end{table}

We adapt an existing framework for Speech Emotion Recognition (SER) trained on four datasets containing speech with labelled emotions~\cite{burnwal2020-speechEmotionKaggle}.  Features such as the Mel Frequency Cepstral Coefficients (MFCC), Mel Spectogram, and Chromagram are extracted from the input speech waveform audio. A convolutional neural network is trained to predict the emotion. For robustness, we incorporate a fifth dataset for training; The Interactive Emotional Dyadic Motion Capture (IEMOCAP)~\cite{busso2008iemocap} dataset containing over 12 hours of audio-visual data of people acting along with emotions annotated. Not all of the datasets contained the same set of emotion labels, and some datasets used different terminology for similar emotions. We summarized how we dealt with these emotional label correspondences in Table~\ref{tab:emotions}.

We guide the painting to have an emotional appearance with the second term of the loss function detailed in Eq.~\ref{eq:loss_speech}. An image-to-emotion prediction model~\cite{szymczak2022emotionPrediction} ($E_{img}$) trained on the ArtEmis dataset~\cite{achlioptas2021artemis} which contains 80,000 artworks with labeled emotions predicts the emotion probability vector of the simulated painting, $e_p$. The emotion from the input speech audio is extracted using our adapted Speech Emotion Recognition model, $E_{speech}(a)$. The cosine distance between these features forms the emotional guidance loss.  

\subsection{Multi-Modal Guidance}

FRIDA makes painting plans by optimizing the brush stroke parameters to achieve an objective which is made up of the weighted sum of loss functions. Eq.~\ref{eq:objective} displays FRIDA's objective function with our two losses $L_{NS}$ and $L_S$ included for natural sound and speech guidance, respectively.

\begin{equation}
\begin{aligned}
&\ddot{p}=\min _p \left[ w_{a}L_{\{NS|S\}}(p, a) +  \sum_{i=1}^4\left(w_i l_i\right) \right] \label{eq:objective}
\end{aligned}
\end{equation}

FRIDA's base model introduces 4 other loss functions, $l_i$, that connect the paintings to modalities text, images, sketches, and styles.
The full objective, Eq.~\ref{eq:objective}, is to  find the painting plan, $p$, that minimizes the weighted sum of losses, given the weights for each loss function, $w_i$ and $w_a$. The optimal painting plan $\ddot{p}$ is the plan that minimizes the weighted sum of loss functions.

\begin{figure*}[t!]
    \centering
    
    \small
    \includegraphics[width=\textwidth,height=6cm]{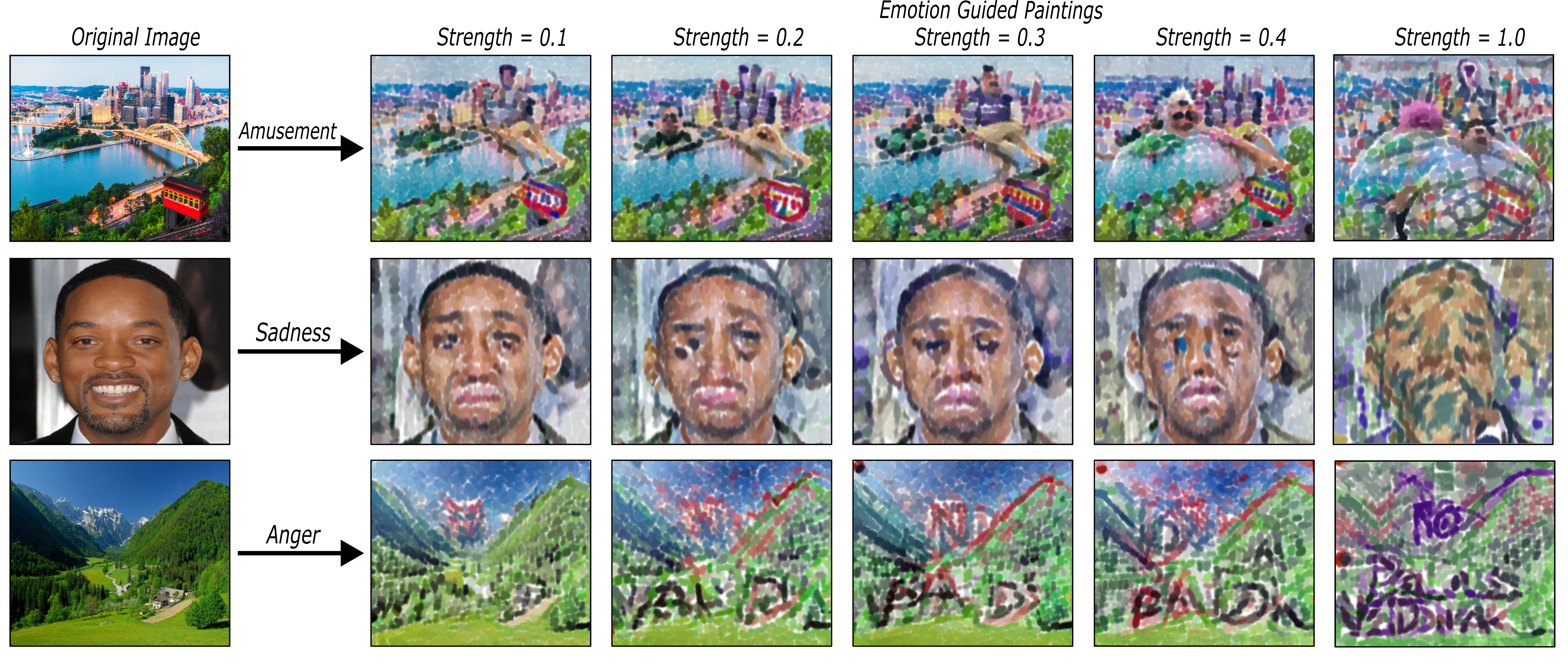}\vspace{-7pt}
    \caption{An image grid depicting the results of emotion guidance with varying guidance strengths. From left to right, each column showcases the progression of the synthesized paintings, with increasing levels of emotion guidance strength, as indicated by the numbers.}\vspace{-0.5cm}%
    \label{fig:emotion_grid}%
\end{figure*}

Fig.~\ref{fig:emotion_grid} shows a gradual progression of the influence of emotion guidance as its loss value weight is increased.

\begin{table}[t]
\centering
\small
\begin{tabular}{rcccccc}
\textbf{}    & \rotatebox[origin=l]{90}{Drill} & \rotatebox[origin=l]{90}{Explosion} & \rotatebox[origin=l]{90}{Fire} & \rotatebox[origin=l]{90}{Raining} & \rotatebox[origin=l]{90}{Thunder} & \rotatebox[origin=l]{90}{Thunderstorm} \\
Drill        & \textbf{3} & 1 & 0 & 0 & 0 & 1  \\
Explosion    & 0 & \textbf{2} & 1 & 0 & 2 & 0  \\
Fire         & 0 & 0 & \textbf{2} & 1 & 2 & 0  \\
Raining      & 0 & 1 & 0 & \textbf{1} & 1 & 2  \\
Thunder      & 2 & 0 & 0 & 2 & \textbf{0} & 1  \\
Thunderstorm & 0 & 0 & 0 & 0 & 0 & \textbf{5} 
\end{tabular}
\caption{The confusion matrix from our survey where participants listened to audio then decided which of the six paintings in Fig.~\ref{fig:just_audio} was generated using that audio. Rows display true labels and columns show the predicted labels.}
\label{tab:confusion_matrix}
\end{table}


\section{Results}
\begin{figure}[t!]
    \centering
    \includegraphics[width=\columnwidth]{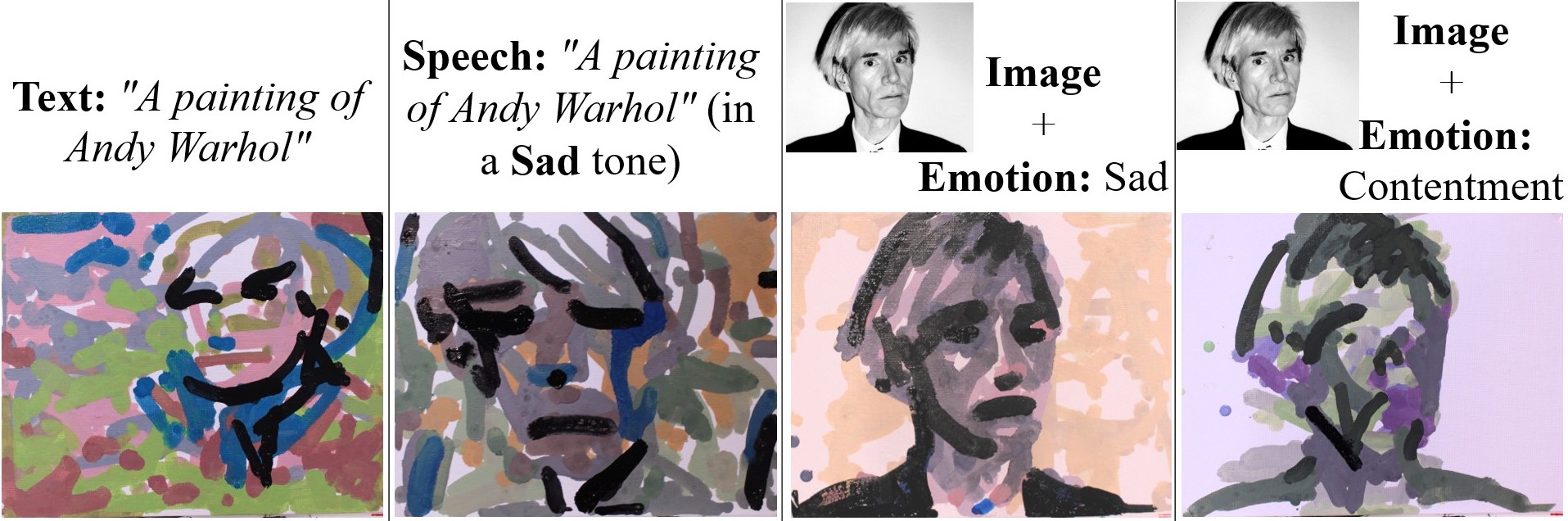}
    \caption{Examples showing how emotion included in the inputs impacts the overall impressions of the paintings. The figure shows the inputs (first row), and the real paintings drawn by \synesthesia{} (second row).}%
    \label{fig:audio_text}
\end{figure}

We demonstrate how natural sound or speech can be used to produce paintings that match the semantic context of the input sound and that align with the user's emotional context. We mainly use simulated paintings for evaluation because \synesthesia{}'s real world paintings take hours to produce, executing hundreds of brush strokes in sequence analogous to human painting. We, however, note that 
\synesthesia{}'s real paintings closely match their simulated counterparts as shown in Fig.~\ref{fig:approach}
due to its Real2Sim2Real technique~\cite{frida2022}. Real painting results are showcased in Fig.~\ref{fig:main_fig} and Fig.~\ref{fig:audio_text}.

\begin{figure}[t]
    \centering
    \includegraphics[width=\columnwidth]{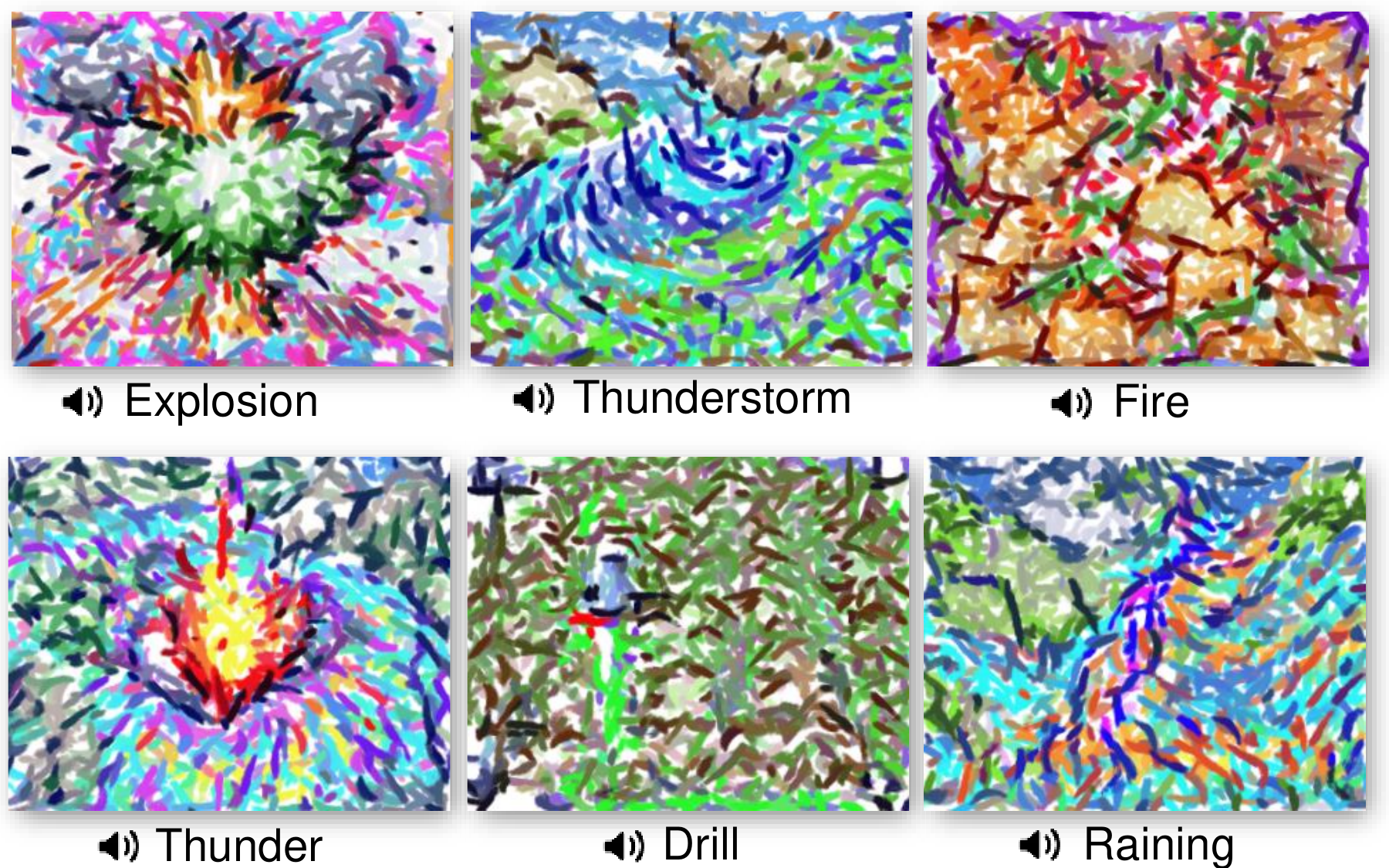}
    \caption{Paintings generated using natural sounds as input.}
    \label{fig:just_audio}
\end{figure}

\begin{figure*}[t!]
    \centering
    \small
    \includegraphics[width=\textwidth]{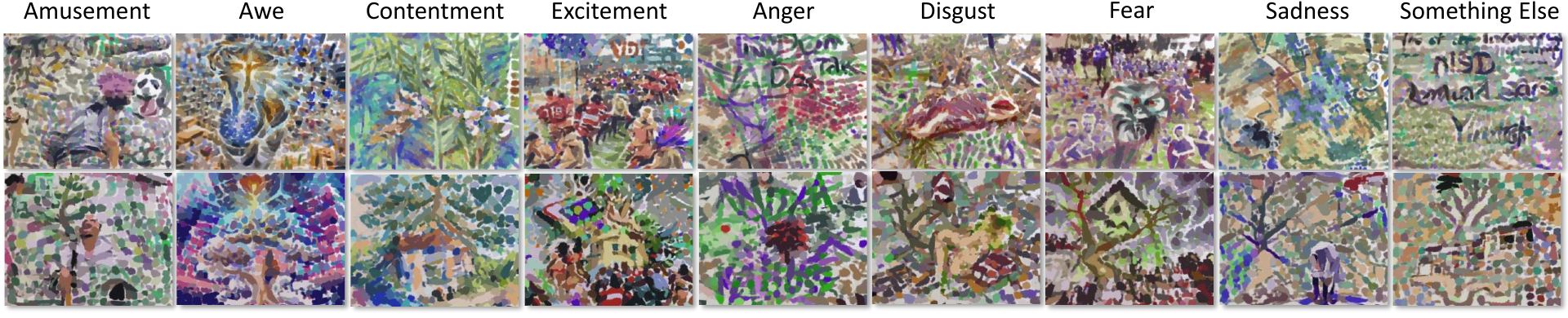}
    \caption{First row images: Painting with only emotion guidance. Second row: painting with both emotion and text ``A house and a tree.''}%
    \label{fig:pure_emotion}\vspace{-10pt}%
\end{figure*}

\begin{figure}[t]
    \centering
    \small
    \includegraphics[width=\columnwidth]{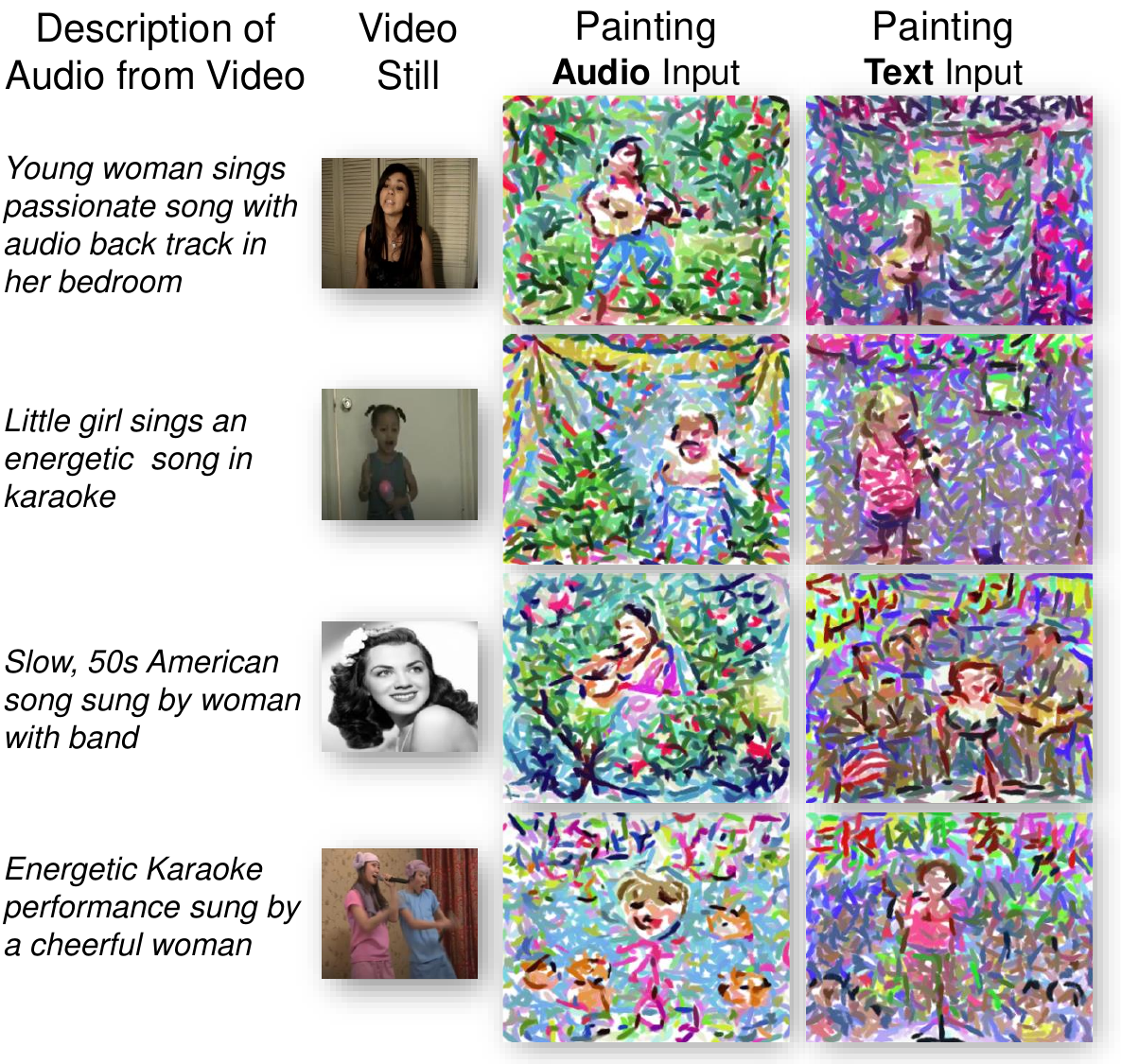}
    \caption{Painting various examples from the VGG-Sound~\protect\cite{chen2020vggsound} categorized as ``female singing" using audio versus using a description of the audio. A video still is shown here for clarity but is not used in image generation.   
    }
    \label{fig:female_singing}
\end{figure}

\subsection{Natural Sounds Guidance}

Fig.~\ref{fig:just_audio} shows that, using only audio as guidance, our approach can generate paintings that capture various natural sounds.
We conducted a survey to quantitatively evaluate the quality of generated images in Fig.~\ref{fig:just_audio} in terms of the human-evaluated semantic similarity between the input audio and the generated painting. 
In the survey, participants were instructed to listen to an audio sample, then select which of the six images (Fig.~\ref{fig:just_audio}) were created using that audio as input.
A random selection would result in 16.7\% accuracy as there were six options; however, participants selected the correct painting 43.3\% of the time.  We conducted the survey through Amazon Mechanical Turk, recruiting 28 participants. Each audio was evaluated by five different participants. The confusion matrix can be seen in Table~\ref{tab:confusion_matrix}. Many of the sounds were similar and this can be reflected with the frequent confusion of thunder, thunderstorm, and raining sounds.


\begin{figure}[ht]
    \centering \small
    \includegraphics[width=0.9\columnwidth]{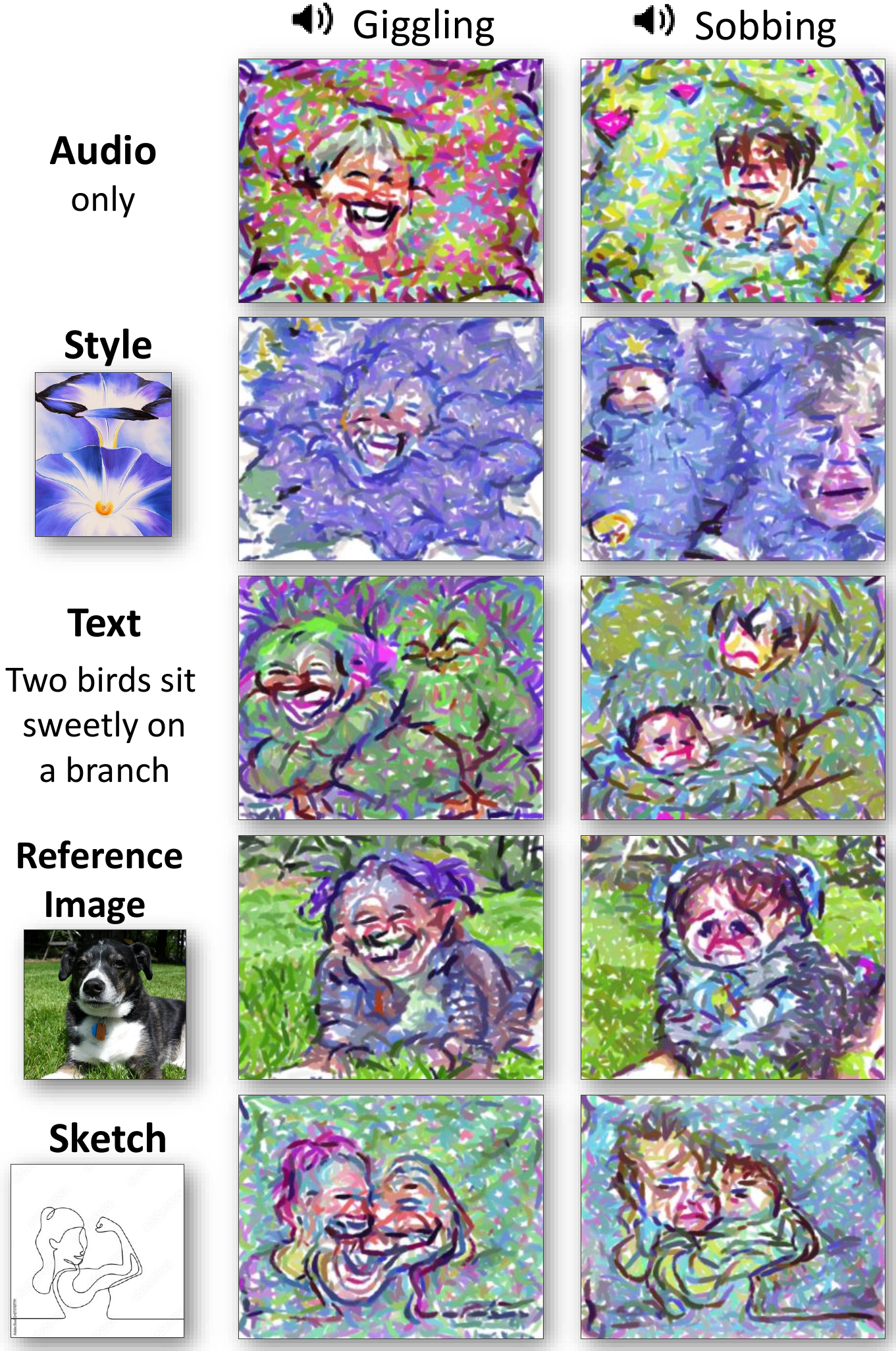}
    \caption{Combining audio as an input modality with other modalities that FRIDA can handle.}%
    \label{fig:audio_and_other_modes}%
\end{figure}

\subsection{Emotion Guidance}
We first present the quantitative evaluation on emotion guided painting generation, followed by real world paintings as qualitative results. 

In this experiment, we examine the quality of generated paintings guided by emotion as a direct input. We use the 9 emotions from the ArtEmis dataset. The first row of Fig.~\ref{fig:pure_emotion} shows \synesthesia{}'s emotional guidance in isolation where the type of emotion was the only input to produce each painting. 
The second row of Fig.~\ref{fig:pure_emotion} shows the paintings generated for the same types of emotion jointly with an additional text input, ``a house and a tree.''; this result shows the emotional influence on the paintings when their desired contents are described by the users. 

Although recognizable content is not generally present in both sets of images, our user survey indicated that the users were able to identify the correct input emotion significantly better when compared to a random guess. 
In a Mechanical Turk study, we presented participants with one of the eight emotions in Fig.~\ref{fig:pure_emotion} (omitting ``Something Else") in writing along with randomized order of paintings generated using the eight emotions. Participants were asked to identify which painting looks most like the given emotion. The 75 participants were correct 26.5\% of the time (random guess would be 12.5\%).
The most common correctly identified emotion was Awe at 60\% correct and the most commonly confused emotions were Amusement incorrectly identified as Excitement 80\% of the time.

Fig.~\ref{fig:audio_text} and Fig.~\ref{fig:main_fig} (upper) present real world paintings illustrating how emotion--as either a direct input or indirect input as part of speech--affects the impression of the resulting paintings. The first column shows the painting of Andy Warhol using a text input only. The second column shows the painting given the same language description but in a sad toned speech. The third and forth columns show the paintings given a photograph of Andy Warhol as input together with two specific types of emotion, sad and contentment. All paintings here were painted using acrylic paints in $8''\times10''$ canvas.




\subsection{Ineffability of Sound}

Sound is nuanced and challenging to describe accurately with language. In this experiment, we qualitatively compare the paintings generated directly from audio with those generated using textual descriptions of the audio as shown in Fig.~\ref{fig:female_singing}. Specifically, we first sampled multiple examples from the held-out set of the VGG Sound~\cite{chen2020vggsound} dataset that were all categorized as ``female singing." 
Given each sample, we generated two paintings: one using the audio as input and the other using the corresponding text description of the audio. While both paintings from audio and text tend to align with the content of the video's sound, they are vastly different from each other. This result hints some evidence that it is challenging to use language to describe the semantic meanings of sounds fully.

\begin{figure}[ht]
    \centering
    \includegraphics[width=\columnwidth]{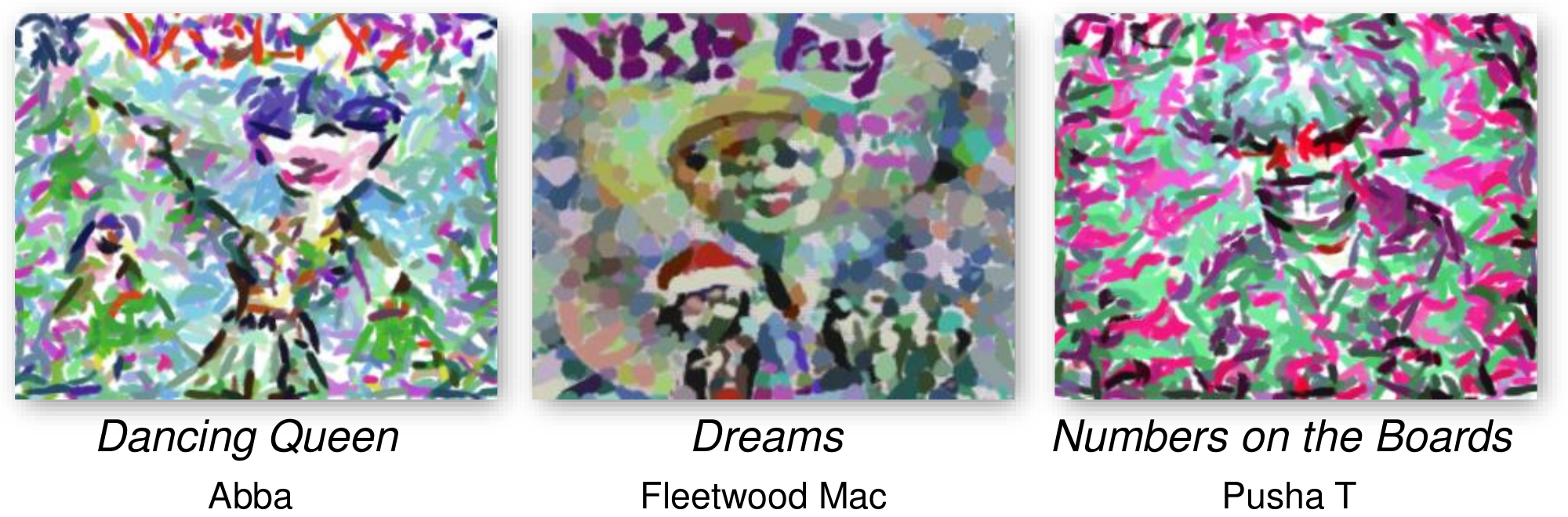}
    \caption{Painting using various Pop songs as input. Genres of the songs from left to right are Disco, Traditional American Folk, and Rap/Hip-Hop}\vspace{-0.5cm}%
    \label{fig:frida_music}%
\end{figure}

\subsection{Additional Results}

\noindent\textbf{Multimodal Guidance}: 
Paintings were generated with audio inputs paired with FRIDA's existing modalities in Fig.~\ref{fig:audio_and_other_modes}.  Loss function weights were adjusted to allow the appearance of both modalities to become prominent.  The results appear abstract, although both modalities are strongly present. 

\vspace{0.1cm}
\noindent\textbf{Music Guidance}:The VGG-Sound dataset used to train our sound encoder contains some music in various forms. Music is a modality that is extremely challenging to represent in any other form such as language or images accurately.  We experiment with the generalization of the image encoder in Fig.~\ref{fig:frida_music}.  The preliminary results are abstract but make promising resemblance to content and general atmosphere of the given songs.

\section{Discussion}

This work generally contributes toward creative human-robot interaction research. We believe that supporting diverse and intuitive input modalities would endow the users an increased control of a robotic painting system. The users, in turn, would feel more ownership over the artwork that they create collaboratively with the system. 

Sound can act as an accessible input modality for people with visual impairments who want to paint.  In future work, we hope to engage with different communities of people who face physical barriers to performing painting  to enable more people to express themselves via the visual art of painting.

\subsection{Limitations}

The generalization of the content that our approach can produce depends on the training data. Our image-to-emotion prediction model is trained on the ArtEmis dataset~\cite{achlioptas2021artemis} which pairs paintings with their labelled emotions. Because art data was used, the model is biased towards guiding our images towards this distribution and away from other image distributions such as photographs. While the VGG-Sounds~\cite{chen2020vggsound} is large with more than 300 categories, it is still not completely general and we observe its limitations in the examples such as music guided paintings.

As in other generative art tasks, evaluating the quality of produced outputs remains challenging.

\synesthesia{}'s paintings appear abstract and noisy compared to recent, pixel-generating models, e.g., Stable Diffusion~\cite{rombach2022stableDiffusion}.  This is partially due to using CLIP and other neural networks that were trained for image categorization rather than generation, as seen with other methods that use CLIP for image generation~\cite{frans2021-clipdraw, schaldenbrand2022styleclipdraw}. Another source of the abstract appearance is from using simple, brush strokes as image primitives rather than pixels. FRIDA plans with a single brush that is not capable of producing very fine details.

\section{Conclusions}


Our approach, Robot Synesthesia, adds sound inputs to the FRIDA~\cite{frida2022} robotic painting platform. When compared to existing sound-guided image synthesis approaches, ours is more general as it does not rely on constrained pre-trianed image generators, e.g., Style-GAN. In addition, we treat natural sounds and speech separately to unlock the important content and emotion held in spoken language. 
While many existing work focuses on robots following commands, Robot Synesthesia is a rare attempt at a robotic system which can hear a human user, understand their emotions, and assist them to express their ideas in visual art.
By supporting audio-based interaction, our work contributes to make a robotic painting system accessible to broader user groups.



\section*{Acknowledgments}
This work was in part supported by NSF IIS-2112633 and the Technology Innovation Program (20018295, Meta-human: a virtual cooperation platform for specialized industrial services) funded By the Ministry of Trade, Industry \& Energy(MOTIE, Korea).

\bibliographystyle{ieeetr}
\bibliography{IEEEfull}

\end{document}